\documentclass[10pt,twocolumn,letterpaper]{article}

\usepackage[pagenumbers]{cvpr}
\usepackage[dvipsnames]{xcolor}

\usepackage[framemethod=TikZ]{mdframed}
\mdfsetup{roundcorner=10pt}

\usepackage{multirow}
\usepackage{listings}
\usepackage{xcolor}
\definecolor{codegreen}{rgb}{0,0.6,0}
\definecolor{codegray}{rgb}{0.5,0.5,0.5}
\definecolor{codepurple}{rgb}{0.58,0,0.82}
\definecolor{backcolour}{rgb}{0.95,0.95,0.92}
\lstdefinestyle{mystyle}{
  backgroundcolor=\color{backcolour}, commentstyle=\color{codegreen},
  keywordstyle=\color{magenta},
  numberstyle=\tiny\color{codegray},
  stringstyle=\color{codepurple},
  basicstyle=\ttfamily\footnotesize,
  breakatwhitespace=false,         
  breaklines=true,                 
  captionpos=b,                    
  keepspaces=true,                 
  numbers=left,                    
  numbersep=5pt,                  
  showspaces=false,                
  showstringspaces=false,
  showtabs=false,                  
  tabsize=2
}
\usepackage[accsupp]{axessibility} 

\definecolor{cvprblue}{rgb}{0.21,0.49,0.74}
\usepackage{hyperref}
\hypersetup{pagebackref,breaklinks,colorlinks,citecolor=cvprblue}

\def\paperID{4}
\def\confName{CVPR}
\def\confYear{2024}

\title{What Makes Multimodal In-Context Learning Work?}

\author{Folco Bertini Baldassini$^1$\and Mustafa Shukor$^1$ \and Matthieu Cord$^{1,2}$  \and Laure Soulier$^1$ \and Benjamin Piwowarski$^1$\\
$^1$Sorbonne Université, CNRS, ISIR, F-75005 Paris, France\\
$^2$ Valeo.ai, Paris, France
}

\begin{document}
\maketitle
\begin{abstract}
Large Language Models have demonstrated remarkable performance across various tasks, exhibiting the capacity to swiftly acquire new skills, such as through In-Context Learning (ICL) with minimal demonstration examples.  In this work, we present a comprehensive framework for investigating Multimodal ICL (M-ICL) in the context of Large Multimodal Models. We consider the best open-source multimodal models (\eg, IDEFICS, OpenFlamingo) and a wide range of multimodal tasks. Our study unveils several noteworthy findings: (1) M-ICL primarily relies on text-driven mechanisms, showing little to no influence from the image modality. (2) When used with advanced-ICL strategy (like RICES), M-ICL is not better than a simple strategy based on majority voting over context examples. Moreover, we identify several biases and limitations of M-ICL that warrant consideration prior to deployment. Code available at \href{https://gitlab.com/folbaeni/multimodal-icl}{gitlab.com/folbaeni/multimodal-icl}
\end{abstract}    
\begin{figure}[t]
    \includegraphics[width=\linewidth]{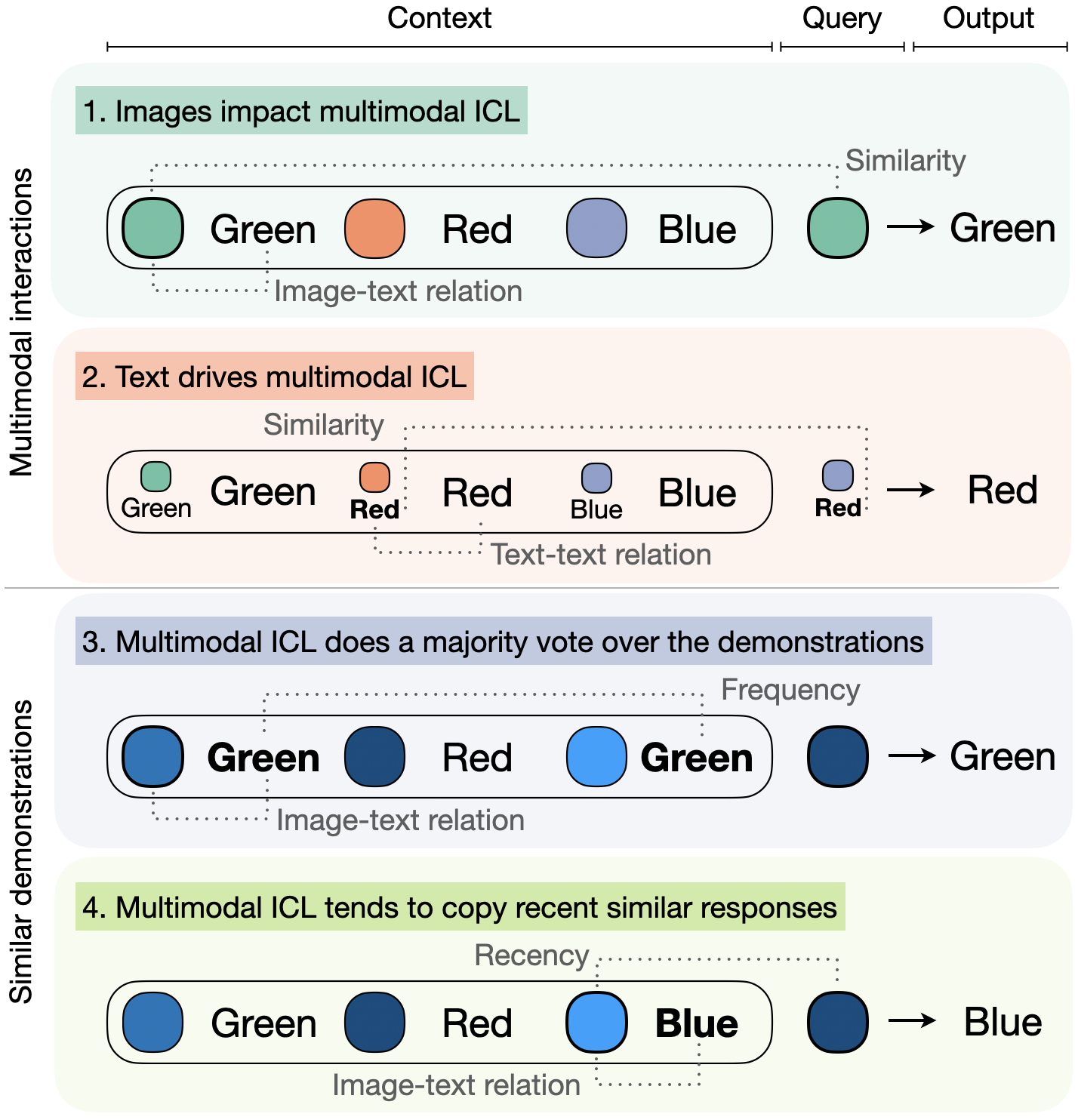}
    \caption{\textbf{Empirical analysis of M-ICL behavior.} 1. Images play a crucial role in image-to-text tasks. 2. M-ICL is mostly driven by text when the task includes both image and text as input.  3. For advanced M-ICL strategies ranking ICL examples by their similarity to the query, the LMM mostly does a majority vote over the demonstration pairs. 4. M-ICL copies the output of the last demonstration pair.}\label{fig:cover}
\end{figure}

\section{Introduction}\label{sec:intro}
Recently, Large Multimodal Models (LMMs) have made considerable progress in comprehending and generating visual and textual content~\cite{li_visualbert_2019, tsimpoukelli_multimodal_2021, alayrac_flamingo_2022, gpt4, gemini}. These models can be seamlessly adapted to solve novel tasks, through In-context learning (ICL)~\cite{brown_language_2020}. It is a training-free approach that consists of augmenting the input prompt with a few pairs (input,output) prepended to the query prompt. This extra context acts as demonstrations that should help the model understand the task at hand. The choice and ordering of examples used in the ICL is decisive to its performance, as observed for retrieval methods~\cite{luo_-context_2024, rubin_learning_2022, liu_what_2021, yu_prophet_2023}, and for multimodal tasks by exploiting CLIP~\cite{radford_learning_2021, yang_empirical_2022,lin_revive_2022}, exemplified by RICES~\cite{alayrac_flamingo_2022}.  
While extensive research has been carried out into conditions and biases of ICL for LLMs~\cite{dong_survey_2022,min_rethinking_2022, zhao_calibrate_2021, liu_what_2021, lu_fantastically_2022}, extending this knowledge to the multimodal domain is not trivial. Besides, multimodal ICL (M-ICL) presents new challenges and biases ~\cite{shukor_beyond_2023, chen_understanding_2023, zhang_out--distribution_2024} that may not be fully addressed by existing unimodal studies. 

In this paper, we propose a comprehensive framework to study M-ICL: using  the best open-source LMM models with M-ICL ability, such as IDEFICS \cite{laurencon_obelics_2023} and OpenFlamingo \cite{awadalla_openflamingo_2023}, we consider a wide range of  multimodal benchmarks that cover Visual Question Answering (VQA), captioning and classification tasks.  To investigate how modalities (image and text) affect the M-ICL behavior, we systematically remove or mix each modality. We then extend our study to approaches that improve ICL with retrieval-based context selection (RICES~\cite{alayrac_flamingo_2022}).

To summarize, we propose a comprehensive framework to evaluate the M-ICL behavior in LMMs.  Our empirical study led to the following findings illustrated in~\Cref{fig:cover}:
    \begin{itemize}
        \item In general, M-ICL is primarily focused on text, overshadowing the role played by images. This is less the case for image captioning and classification tasks.
        \item For advanced similarity-based context selection M-ICL methods, the LMM models behave so far not better than  a majority voting mechanism over the context demonstrations. 
        \item We also identify a major flaw in these advanced similarity-based methods. They suffer from recency bias, where the model tends to "copy" the answer of the last example in context. This sheds light on several limitations that should be considered before deployment. 
    \end{itemize}

\section{Related work}

\paragraph{Multimodal models}
 have undergone significant advancements recently~\cite{zhang_mm-llms_2024}, by moving towards more unified models that can support a myriad of tasks and modalities \cite{wang2022ofa,shukor_unified_2023,lu2022unified,mizrahi20244m,alayrac_flamingo_2022,li_blip-2_2023}. These models are generally built on top of pre-trained LLMs and visual encoders that are simply connected by a linear transformations~\cite{shukor_ep-alm_2023,liu_visual_2023, vallaeys2024improveddepalm,liu_improved_2023, eichenberg_magma_2022, tsimpoukelli_multimodal_2021, wang_simvlm_2022, mokady_clipcap_2021}, or transformer-based mechanisms \cite{li_blip-2_2023,alayrac_flamingo_2022,laurencon_obelics_2023}. The level of performance of these models has started to approach those of LLMs, especially after multimodal instruction tuning~\cite{xu_multiinstruct_2023, liu_visual_2023, dai_instructblip_2023, luo_cheap_2023, li_otter_2023}. In addition, several models can now support ICL \cite{alayrac_flamingo_2022}, arguably due to training on interleaved image-text datasets. In this work, we focus on the best open-source models with ICL abilities (IDEFICS~\cite{laurencon_obelics_2023} and OpenFlamingo~\cite{awadalla_openflamingo_2023}), and in particular, IDEFICS that achieves comparable performance to Flamingo. 

\paragraph{In-Context Learning (ICL)} is a paradigm that allows language models to learn tasks given only a few demonstrations~\cite{brown_language_2020} and is particularly effective for tackling more complex and reasoning-based tasks~\cite{wei_chain--thought_2023,wei_emergent_2022, lu_are_2023}.
To explain ICL, studies compare it with gradient descent~\cite{von_oswald_transformers_2023, akyurek_what_2023, yadlowsky_pretraining_2023, dai_why_2023, von_oswald_transformers_2023, olsson_-context_2022} and examine the inner workings of the models~\cite{olsson_-context_2022, hendel_-context_2023}.
ICL is highly sensitive to the prompt and choice of demonstrations,  \citet{min_rethinking_2022} indicates that the format of the prompt and distribution of the words matter, though the importance of labels is debated~\cite{yoo_ground-truth_2022, wang_label_2023}. Interestingly, ~\cite{pan_what_2023} discusses task recognition and task learning, where the former requires a few examples to understand the task format, and the latter to reproduce the input-output mapping. This depends on multiple factors such as if the model has been instruction tuned~\cite{peng_when_2023}, the model size~\cite{wei_larger_2023, wei_symbol_2023}, and the semantics of the prompt~\cite{webson_prompt-based_2022}, affecting the necessary number of shots. Studies also identify recency and majority biases~\cite{zhao_calibrate_2021} and order sensitivity~\cite{lu_fantastically_2022}.

\paragraph{Multimodal ICL.}
ICL can be extended to multimodal models after training on interleaved image-text datasets ~\cite{alayrac_flamingo_2022, tsimpoukelli_multimodal_2021,tai_link-context_2023, zhao_mmicl_2023,li_otter_2023}.
To further enhance M-ICL, several works try to use better context demonstrations using similarity sampling-based approaches ~\cite{yang_empirical_2022, liu_what_2021, lin_revive_2022, yu_prophet_2023, gui_kat_2022,alayrac_flamingo_2022, chen_understanding_2023}. Despite being effective, especially in handling out-of-distribution tasks ~\cite{zhang_out--distribution_2024}, several works have tried to highlight several flaws. In particular, increasing object hallucinations and the limited ability to solve complex tasks such as instruction following or compositional image-text matching~\cite{shukor_beyond_2023}. In addition, \citet{chen_understanding_2023} study OpenFlamingo
and find that the image plays a marginal role in VQA tasks, raising questions about the effectiveness of ICL in a multimodal context. 

\section{Analysis framework of M-ICL}\label{sec:framework}

For M-ICL, LMMs process inputs composed of a query $Q$ and a context $C$. The query $Q$ includes an image $I$ and an optional associated text $T$, which can be a question, instruction, or additional information. The context $C$ comprises $N$  demonstrations (examples) from the training dataset $D$, each containing images and texts along with their corresponding responses $R$. M-ICL can be written as follows:
\vspace{-1em}
\begin{equation}
    C=\left((I_i,T_i,R_i)\right)_{i\in D_C},\ \ O = \text{LMM}(C, (I_Q, T_Q))
\end{equation}
Our similarity sampling method is RICES~\cite{alayrac_flamingo_2022}. 
Given a query $Q$, it retrieves the $N$ most similar demonstrations from the training set according to \( S_{iq} = s(I_i, I_Q) + s(T_i, T_Q) \), where $s$ represents the similarity score calculated by the visual encoder CLIP~\cite{radford_learning_2021}. These demonstrations are arranged in the context in order of increasing similarity.

\subsection{Research questions \& analysis methodology}\label{sec:methodology}
Our objective is to understand how different modalities affect M-ICL -- here text and image. While there are several methods for demonstration retrieval in the literature~\cite{gao_retrieval-augmented_2024,liu_what_2021, lin_revive_2022, yu_prophet_2023, gui_kat_2022, rubin_learning_2022}, there's limited work~\cite{alayrac_flamingo_2022, yang_empirical_2022} for M-ICL and consequently little analysis of these methods. We believe that it is essential to investigate how ICL's sensitivity factors apply to these methods and identify their limitations. We address the following research questions:

\textbf{RQ1: How does each modality influence M-ICL?} To analyze the effect of each modality, we modify the context $C$ by adjusting either $I$ (images) or $T$ (text). We describe the procedure for $I$, but the same method applies to $T$. We either completely remove the image component, resulting in a new context defined as $((\varnothing, T_i, R_i))_{i\in D_C}$, or randomize this modality by using random images from the demonstration dataset. In the later, the altered context is represented as $((I_j, T_i, R_i) | j \neq i)_{i\in D_C}$.  
We also conduct experiments with RICES to identify any behavioral differences. 

\textbf{RQ2: Which kind of shortcuts influence M-ICL?}
We are interested in whether M-ICL involves genuine learning from demonstrations, or if it relies on what we name ``shortcuts''. Using Generalized Linear Models (GLM) and Spearman's rank correlation, we evaluate the relationship between the similarity of the demonstrations to the query and their performance outcomes. We compare random sampling with RICES to understand M-ICL's behavior, focusing on the improvements attributed to RICES. This analysis aims to understand the reason of these improvements and whether they reveal any emerging behaviors that suggest reliance on shortcuts.
We then turn to the question of what performance gains can be attributed to RICES or to the m-ICL of LLMs. 
More precisely, for classification tasks, we rely on a simple RICES based KNN where the predicted answer $O'$ is given by $\text{argmax}_R \left( \sum_{\{i \in D_C | R_i = R\}} e^{S_{iq}} \right)$. For generation tasks (VQA and captioning), we also rely on another set of analysis, since the KNN approach is not the most adapted. 
Finally, we investigate another factor impacting M-ICL, namely the recency bias that complement our analysis on the relationship between the similarity of the context answer to the target one.

\subsection{Experimental setup}

\paragraph{Datasets} In our study, we investigate various tasks, including image captioning, classification, and visual question answering. For \textbf{captioning}, we employ the COCO dataset~\cite{chen_microsoft_2015} and the Flickr30k dataset~\cite{plummer_flickr30k_2016}, where each image is annotated with five captions; we select one caption randomly for our experiments and evaluate using the CIDEr~\cite{vedantam_cider_2015} metric. In \textbf{classification}, we use the CIFAR-100~\cite{Krizhevsky2009LearningML} and ImageNet~\cite{russakovsky_imagenet_2015} datasets, with 100 and 1000 classes, respectively. The predicted class is the one whose label has the smallest Levenshtein distance to the model's output. We use accuracy as the metric. An alternative would be to instruct the model to choose among all the classes, but this has a high computational cost.
We also examine the Hateful Memes~\cite{kiela_hateful_2021} and Rendered SST2~\cite{openai_clip_2024, SocherEtAl2013:RNTN} datasets for detecting hate speech and performing sentiment analysis through OCR, measuring performance by exact match accuracy. For\textbf{ visual question answering}, we use the VizWiz~\cite{gurari_vizwiz_2018}, VQAv2~\cite{agrawal_vqa_2016}, OK-VQA~\cite{schwenk_-okvqa_2022}, TextVQA~\cite{singh_towards_2019}, ScienceQA~\cite{lu_learn_2022} (only items containing images), and MMMU~\cite{yue_mmmu_2023} datasets, 
covering a range of applications from assisting visually impaired users to requiring scientific reasoning, 
with VQA accuracy as metrics for most, except accuracy for multiple-choice formats for ScienceQA and MMMU. The test set is composed of a maximum of 5000 items, chosen randomly if the original dataset exceeds this number. This set remains the same across all tests, serving as a consistent basis for comparison. Additionally, the entire training dataset is used as the support set for M-ICL demonstrations.

\begin{figure*}[ht]
    \centering
    \begin{subfigure}[t]{0.32\linewidth}
        \includegraphics[width=\linewidth]{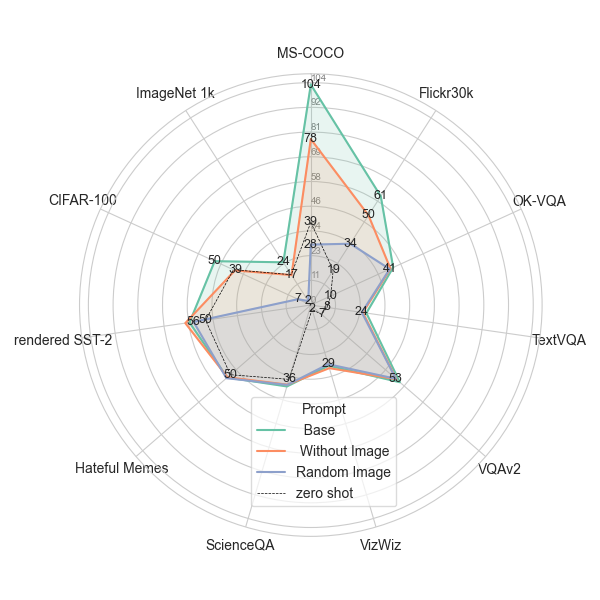}
        \caption{Altering image - 16 shots}
        \label{fig:radar_image}
    \end{subfigure}
    \hfill
    \begin{subfigure}[t]{0.32\linewidth}
        \includegraphics[width=\linewidth]{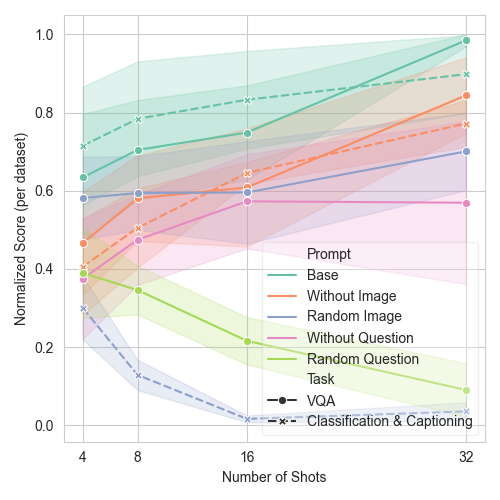}
        \caption{Performance vs number of demonstrations. 
        }
        \label{fig:normalized}
    \end{subfigure}
    \hfill
    \begin{subfigure}[t]{0.32\linewidth}
        \includegraphics[width=\linewidth]{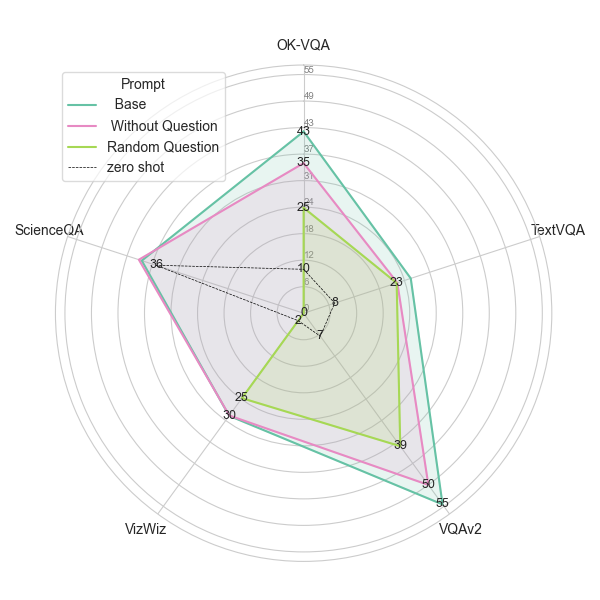}
        \caption{Altering question - 16 shots}
        \label{fig:radar_vqa}
    \end{subfigure}
    \caption{\textbf{Influence of each modality on the M-ICL performance.} We show (a) the 16 shot performances of M-ICL with different contexts: baseline context (green), demonstration without images (orange), or with random images (blue).  For VQA (c), we also consider the case where questions $T$ of the demonstrations are removed (pink), or replaced by a random question (green). In (b), we show the evolution of performance when the number of shots varies.
    }
    \label{fig:combined_figures}
\end{figure*}

\paragraph{Models and ICL details.}
We conduct our tests with IDEFICS~\cite{laurencon_obelics_2023}  9B version (for OpenFlamingo, results are reported in the Appendix \cref{par:openflamingo}). For RICES we use the CLIP version "openai/clip-vit-large-patch14". 
Unless specified, demonstrations are chosen randomly.
For captioning and classification tasks (image-to-text tasks), the demonstrations consists of interleaved image and captions/classes. For VQA datasets, the text consists of the question-answer pairs.
We do not use explicit task instruction, letting the model understand the task from its context. 
We repeat each experiment 3 times and report the averaged results.

\section{RQ1: How does each modality influence M-ICL?}\label{sec:text_image}
In this section, we try to answer RQ1, i.e we investigate the influence of each modality on M-ICL and their interactions by manipulating the context (text or images).
We conduct our study with randomly sampled demonstrations and extend to the retrieval M-ICL such as RICES in \Cref{sec:rices_analysis}.
We summarise the results in \Cref{fig:combined_figures}, presenting the scores for the 16-shot scenario with both contexts of altered images (\Cref{fig:radar_image}) and texts (\Cref{fig:radar_vqa}). Additionally, we illustrate the effect of the number of demonstrations in \Cref{fig:normalized}. To make values comparable across tasks, we normalize the measures.

\subsection{Images impact M-ICL}\label{sec:no_image}
In \Cref{fig:radar_image}, we observe that image-to-text tasks like captioning and classification are highly affected when altering the images. Compared to the context baseline that consists of images and their correct classes/captions, using random images or removing them from the context leads to a significant drop in performance. The  performance for datasets such as CIFAR and ImageNet is close to the level of a zero-shot m-ICL, and for MS-COCO it is even worse. This phenomenon is corroborated in \Cref{fig:normalized}, where we show that adding more demonstrations with random images has a strong negative impact on image-to-text tasks, in stark contrast to the initial demonstrations.

\begin{figure}[t]
    \centering
    \includegraphics[width=\linewidth]{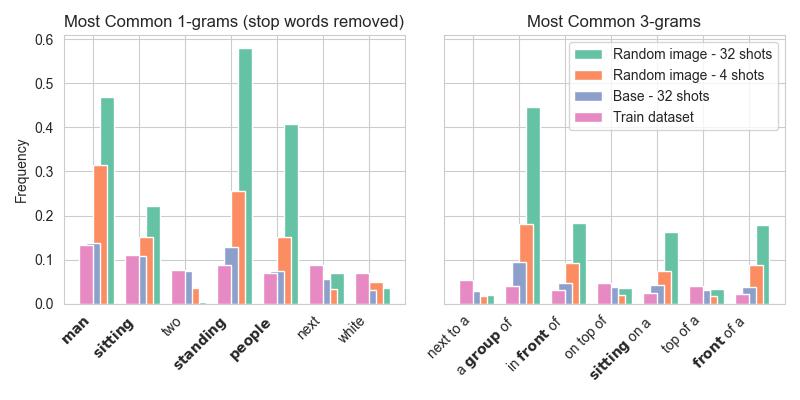}
    \caption{
    \textbf{M-ICL tends to output the most frequent words of the context.} We show the frequency of the most common words (excluding stop words) and 3-grams in the COCO dataset, which is used to construct the context demonstrations. We comprare the words frequency of the model outputs, with normal  (blue) and random images (orange and green), to the dataset words frequency (pink).
    }
    \label{fig:idefics_coco_bigrams}
\end{figure}

To understand the effect of using random images, in \Cref{fig:idefics_coco_bigrams} we examine the model's output in this setup. Our analysis focuses on the most common n-grams found in the captions over the whole dataset (pink) within the context, looking at their frequency in the model's output. We compare the base prompt (blue) when 32 demonstrations are used, against random images setup in 4 shot (orange) and 32 shot (green). In the case of the base prompt, the distribution appears similar to that of the context, indicating a similar input and output distribution of words. However, in scenarios involving random images, there is a noticeable shift towards an over-representation of the most frequent n-grams, and the more demonstrations the more this happens. This suggests that the mismatch between images and their corresponding textual outputs in the demonstrations causes the model to switch to a generic mode, in which it tends to output the most frequent words in the dataset used to construct the ICL context. 

These results suggest that \emph{demonstration images influence} the performance of M-ICL in image-to-text tasks, and that the model leverages the relationship between visual inputs and textual outputs.
We discuss the potential reasons for this behavior in section \ref{sec:understanding}.

We now turn to VQA which exhibits a different behavior. Altering or omitting images results in a minor decrease in performance, typically between 1.2 to 1.5 points from the base prompt (\Cref{fig:radar_image}). This suggests that the inclusion of textual information (i.e. questions) diminishes the model's dependence on visual data, a topic we explore in the next section.

\subsection{Text drives M-ICL}\label{sec:no_image_vqa}
In VQA, which has both image and text (i.e. questions) as input,  \Cref{fig:radar_vqa} illustrates that removing the question (purple) results in an average drop of 3.5 points. Moreover, replacing it with a random question (green) leads to an average decrease of 9.5 points. We further observe (\Cref{fig:normalized}) that the decrease worsens with an increasing number of shots\footnote{In practice, M-ICL often outputs 'no', the most frequent answer in the dataset}. 

For text-to-image tasks, \Cref{fig:radar_image} provides also insights into the role of text, as scenarios without images (orange) correspond to a scenario with only text. In classification tasks, where the text has limited information, i.e. just the one-two word labels,  the text-only scenario performs as poorly as the zero-shot setup (black), with only a 0.47\% increase in accuracy. However, in captioning, where the text is richer, M-ICL enables capturing the style of the captions and/or the distribution of words, resulting in an average improvement of 31 points over the zero-shot approach.
These results indicate that text influences M-ICL when it carries sufficient semantic content.

In summary, in classification tasks, text has a minor impact compared to images. When text becomes richer, particularly in the context of captioning, the use of text alone can improve zero-shot methods by 31 points. Incorporating images further enhances performance by an additional 20 points, underscoring the importance of both modalities. 
In tasks like VQA, textual information becomes dominant and significantly influences performance, with random text  leading to a significant drop in performance. We conclude that while images do have an impact on M-ICL, \emph{textual information takes precedence and drives} the model's decision-making process.

\subsection{How do retrieving similar demonstrations affect interactions?} \label{sec:rices_analysis}
\begin{figure}[t]
    \centering
    \includegraphics[width=\linewidth]{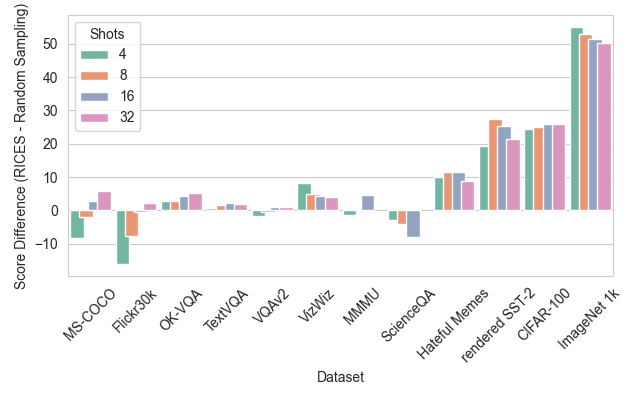}
    \vspace{-1em}
    \caption{
    \textbf{RICES improves M-ICL performances on most datasets.}  
   Score differences between RICES and random sampling, with a varying number of demonstrations and across various datasets, with their respective metrics.}\label{fig:rices_image}
\end{figure}

In the previous section, demonstrations were randomly sampled. Here, we turn to similarity-based (RICES) M-ICL, and analyze which observations still hold and which don't.
First, \Cref{fig:rices_image} shows that in most cases RICES leads to better performance. For captioning tasks, the more demonstrations, the better the performance. For VQA, the use of RICES leads to improvements of between 2 and 5\% for most datasets. The most significant improvements are in classification, where gains range from 10 to 50\%. 

Investigating the factors influencing these improvements and how each modality contributes can help us understand better multimodal interactions. We follow the same procedure as with random sampling: We investigate the effect of disrupting the alignment between visual and textual parts, while maintaining one modality closely related to the query. Additionally, we explore which modality is pivotal for the improvements by computing similarity based on different modality choices.

\begin{figure}[t]
    \centering
    \vspace{-1em}
    \includegraphics[width=0.8\linewidth]{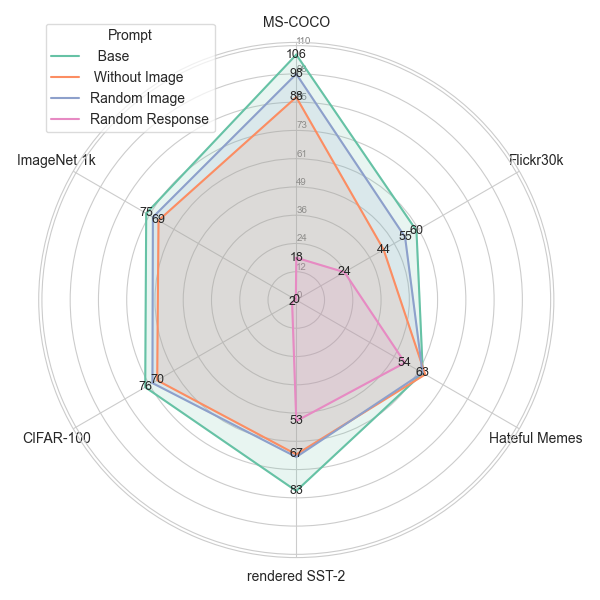}
    \vspace{-1em}
    \caption{
     \textbf{Influence of each modality on RICES M-ICL performance.} 
    We show the 16 shot performances of RICES M-ICL with different contexts: baseline prompt (green), demonstrations without images (in orange), random images paired with responses from demonstrations sampled using RICES (in blue), and random responses paired with images from demonstrations sampled using RICES (purple). }\label{fig:rices_no_image}
\end{figure}

\paragraph{Disrupting image-text alignment}  
In \Cref{fig:rices_no_image} we observe that there is no significant degradation when removing the images or replacing them with random ones. The context with random images (in blue), where only the demonstration responses resemble the query, yields results comparable to random sampling and is slightly better than removing images.
Furthermore there is no noticeable drop in performance as the number of examples increases, which is different than when using randomly sampled demonstrations (as shown in Appendix \cref{fig:rices_no_modality_full}).
On the other hand, random responses (in purple) show a significant decrease in performance (i.e. only the demonstration images are similar to the query's). 

In particular, when substituting the responses in the context by  random ones, the drop is more important in RICES than with random sampling (\eg, as shown in Appendix \Cref{fig:rices_no_modality_full}; for random sampling and image-to-text tasks, random image and random label is equivalent). 
Having the wrong responses for images similar to the query, might push the model to naturally output the wrong response as well. 

Overall, the results above suggests that images serve as a prior for the demonstrations, which is confirmed in the analysis conducted in \cref{sec:understanding}.

\begin{figure}[t]
    \centering
    \includegraphics[width=\linewidth]{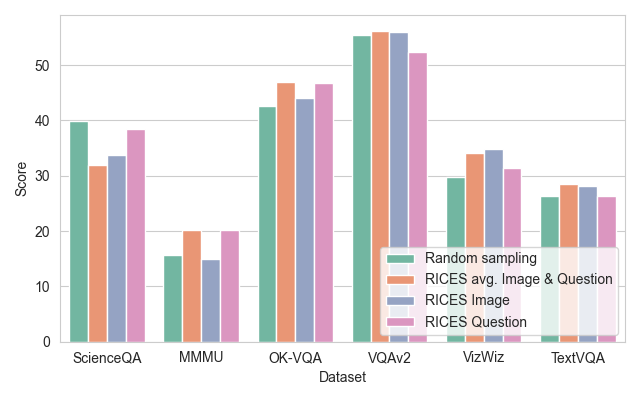}
    \vspace{-1em}
    \caption{\textbf{Influence of different similarity metrics on RICES M-ICL performance.} We show the performances of M-ICL with various sampling methods: Random (in green), RICES (in orange), RICES based only image similarity (in blue), and RICES based only on question similarity (in pink).}\label{fig:rices_vqa}
\end{figure}

\paragraph{Retrieving demonstrations similar to text or image query?}\label{sec:rices_key}
In the case of VQA, the question is composed of text and images. As described in \Cref{sec:framework}, $S_{iq}$ is the sum of CLIP textual and visual similarities. In \Cref{fig:rices_vqa}, to further explore the effect of each modality, we compare this baseline (orange) to using only CLIP image similarity (blue) or CLIP text similarity (pink).
Results vary across different datasets, however for TextVQA, VQAv2, and VizWiz, using image similarity has a better outcome, while  textual similarity is better for MMMU, OK-VQA, and ScienceQA. 
This might be explained by the nature of each dataset: TextVQA, VQAv2, and VizWiz necessitate images for accurate responses, whereas MMMU, OK-VQA, and ScienceQA are more dependent on textual information. To conclude, using the right similarity highly depends on the actual dataset, and there is no clear indication of which to choose for M-ICL models.

\section{RQ2: Which kind of shortcuts influence M-ICL?}\label{sec:sampling}\label{sec:understanding}

In this section, we answer to RQ2, i.e. we try to explain the M-ICL behavior with random or similarity-based demonstrations. More precisely, we investigate whether M-ICL performance can be partially explained by the fact that the demonstration responses can be close to the desired response, and the M-ICL model do a ``soft copy'' of the demonstration responses. Formally, we hypothesize (1) that, given a demonstration $i$ and a query $q$ the similarity function $S_{iq}$ has a correlation with the CLIP score between the responses $R_i$ and $R_q$, denoted $S_{iq}^R$, i.e. if demonstrations inputs are similar to the query inputs, the same applies for the responses. Furthermore, we also hypothesize (2) that, given a context $C$ composed by demonstrations $i$, the average of the similarities $S^R_{iq}$ of the demonstrations responses $R_i$ with the target response $R_q$ correlates with performances, i.e., the closest the context responses to the target one, the better the generated one.

To verify these hypotheses, we compute both General Linear Model (GLM) coefficients and Spearman correlation to characterize the relationship between different factors described above.
In the first column of \Cref{tab:corr}, we compare $S_{iq}$ and $S^R_{iq}=s(R_i, R_q)$ with $s$ the CLIP similarity across all demonstrations (hypothesis 1). With RICES, we can observe a positive Spearman correlation, especially for classification (SST2) and text-to-image (COCO) datasets, slightly less for VQA (VQAv2). The regression coefficient, close to 1, shows that the similarities almost match in average. We also observe that correlation drops when using random samples, showing that this relation holds only when looking at more similar demonstrations.
In the second column of \Cref{tab:corr}, we look at the relationship between (a) the average similarity $avg(S^R_{iq})$ between a demonstration and target response; and (b) the performance of M-ICL. We again only observe correlation in all cases when using RICES demonstrations.

\begin{table}[t]
\small
\centering
\begin{tabular}{ll|cc|cc}
\toprule
\multirow{2}{*}{Dataset} & \multirow{2}{*}{Sampling} & \multicolumn{2}{c|}{$S_{iq} \sim S^R_{iq}$} & \multicolumn{2}{c}{$avg(S^R_{iq}) \sim $ score} \\
& & GLM & Sp. & GLM & Sp. \\
\midrule
\multirow{2}{*}{COCO} & Random & 0.69 & 0.16 & 0.96 & -0.01 \\ 
& RICES & 0.75 & 0.37 & 0.51 & 0.22 \\ 
\midrule
\multirow{2}{*}{VQAv2}& Random & 1.33 & 0.10 & 0.66 & 0.25 \\
& RICES & 1.01 & 0.18 & 0.61 & 0.33 \\ 
\midrule
\multirow{2}{*}{R. SST2}& Random & 0.80 & 0.05 & 1.01 & 0.22 \\
& RICES & 0.89 & 0.29 & 0.96 & 0.35\\
\bottomrule
\end{tabular}

\caption{
\textbf{Correlation between input and output similarities and performance.}
The correlation between inputs and outputs of any given demonstration and any query is represented by \(S_{iq} \sim S^R_{iq}\). Here, \(S_{iq}\) refers to the similarity of the inputs of the demonstration and the query, while \(S^R_{iq}\) refers to the similarity of their responses. \(avg(S^R_{iq}) \sim \text{score}\) represents the correlation, for a given set of demonstrations $i$ within a context $C$, between the mean similarity of the demonstration responses with the query's
and the overall score.
We show the coefficients of the Generalized Linear Model (GLM) as well as Spearman's rank correlation (Sp.), with all p-values $< 0.01$}\vspace{-1em}
\label{tab:corr}
\end{table}

These observation support our initial explanation of M-ICL performance in the case of RICES (this is less clear otherwise), i.e. RICES is effective because it retrieves responses that closely match the target one.
This raises the question of whether the performance gains from M-ICL are simply due to better context responses acting as \emph{shortcuts}, or whether there is genuine learning involved, with demonstrations that are more similar to the query proving to be more useful. In what follows, we study two potential shortcuts: one being that M-ICL might simply exploit the presence of more accurate or relevant responses in the context, and the other being that the most similar demonstrations, whose response is probably the same or close to the query's, are the most recent, and the model could be leveraging this recency. The remaining of this section aims to explore and clarify these two possibilities.

\subsection{M-ICL does a majority vote over the demonstrations}\label{sec:knn}
We dive into the first possibility, which examines the impact of having more accurate or relevant responses in the context. We aim to assess the effectiveness of M-ICL with demonstrations similar to the query by comparing the performance of M-ICL and a simple \textit{RICES KNN} baseline described in  \cref{sec:methodology}.

\begin{figure}[t]
    \centering
    \vspace{-2em}
    \includegraphics[width=0.8\linewidth]{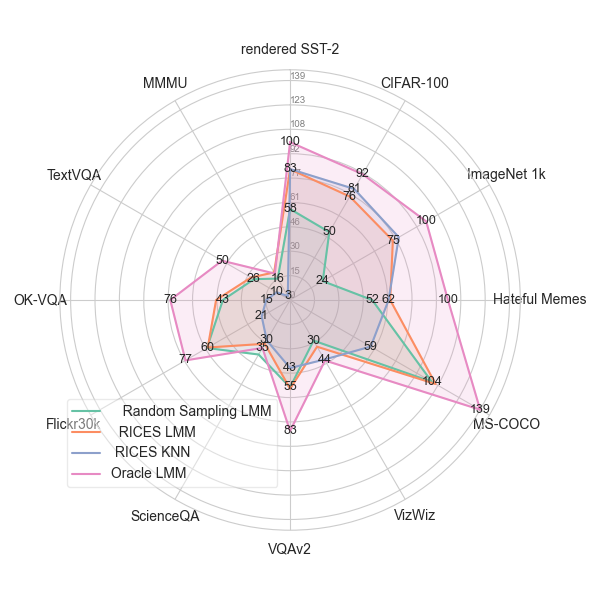}
  \vspace{-2em} 
  \caption{\textbf{M-ICL comparison with majority voting.} We show the 16 shot performances of M-ICL with random sampling (green), M-ICL with RICES (orange), and RICES KNN (blue), M-ICL with RICES using oracle response as similarity (pink).}\label{fig:rices_knn}
\end{figure}

\Cref{fig:rices_knn} illustrates that for classification, RICES KNN (blue) obtains similar performances than M-ICL when using the same demonstration (orange), and outperforms the random sampling setup (green). 
In particular RICES M-ICL struggles to surpass RICES KNN, and this is particularly visible for SST-2, where increasing the number of demonstrations decreases the performances for both the KNN and ICL (see Appendix~\ref{fig:rices_key_full}).  

To further show this majority voting effect, we observed that ensuring that the labels are uniformly distributed with the demonstrations degrades the performance of both M-ICL and the KNN (see Appendix~\ref{tab:balanced_sampling}). 
This suggests that M-ICL leverages similar demonstrations by leveraging the distribution of context responses, rather than actually learning. Said otherwise, in classification tasks, M-ICL's effectiveness is comparable to that of a KNN, and M-ICL does not seem to be useful.

\begin{figure*}[t]
    \centering
    \begin{subfigure}{0.48\linewidth}
        \centering
        \includegraphics[width=1\linewidth]{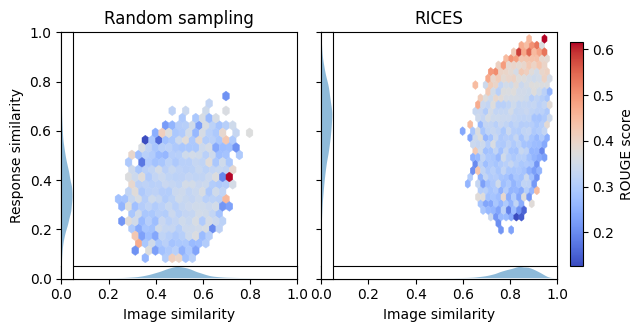}
        \caption{COCO dataset}\label{fig:similarity_coco}
    \end{subfigure}
    \hfill
    \begin{subfigure}{0.48\linewidth}
        \centering
        \includegraphics[width=1\linewidth]{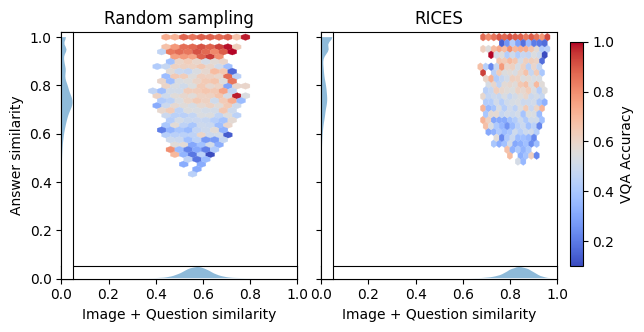}
        \caption{VQA dataset}\label{fig:similarity_vqa}
    \end{subfigure}
    \caption{\textbf{Relation between responses similarities with the performances.} 
We show the 4 shot performances of M-ICL in relation with respectively the input (Image + Question) and response (Answer) similarity of the demonstrations with the query.}
    \label{fig:sim}
\end{figure*}

In open-ended generation tasks, i.e. captioning and visual question answering, majority voting is insufficient. Here the baseline method falls short against random sampling and the RICES approach shows small improvements. \Cref{tab:vqa} and \Cref{fig:sim} show that there is a correlation especially between the responses and performance while this is not the case for the images and texts. This is more true with RICES, but also present for random sampling in VQA. 

To further analyze this phenomenon, we introduce \textit{Oracle} RICES which leverages the similarity metric $S^R_{iq}=s(R_i,R_q)$ that uses the ground truth response $R_q$.  This approach enables us to select examples with responses that closely match the desired answer. In VQA, if "yes" is the correct answer, the chosen examples will all share this answer despite differences in image or text content.
\Cref{fig:rices_knn} illustrates this method in pink and that it significantly outperforms the others methods, providing an upper limit for the RICES approach. 
This in turn show that (a) for open-ended generation m-ICL can do intelligent soft copy when provided close responses; (b) that the used RICES similarity does not select enough demonstrations with a high target response similarity which can improve substantially the performance.

\begin{table}[t]
\centering
\footnotesize
\vspace{-0.5em}
\begin{tabular}{l|l|rrrrrr}
\toprule
Dataset & Sampling & $I$& $T$ &$ R$ & $IT$& $TR$ & $IR$ \\
\midrule
COCO    & Random    & 0.61 & -    & 0.40 & -    & -    & -0.73 \\
        & RICES     & -0.01 & -    & 0.64 & -    & -    & -0.22 \\
\midrule
VQA     & Random    & -0.55 & -0.18 & 0.41 & 0.29 & 0.29 & 0.55 \\
        & RICES     & 0.02  & -1.00 & 0.98 & 0.24 & 0.24 & 0.17 \\
\bottomrule
\end{tabular}

\caption{
\textbf{Influence of demonstration's parts on the performances.}
GLM coefficients (with the score as the response variable) of similarities of context image $I$, text $T$, response $R$  with target ones, as well as their interactions, i.e. Image*Text ($IT$), Text*Response ($TR$), Image*Response ($IR$). For each context, we select the maximum of each value across the demonstrations. All coeff. have a p-value $< 0.001$}
\label{tab:vqa}
\end{table}

\subsection{M-ICL tends to copy recent similar responses}\label{sec:recency}
Another factor impacting the performance can be the ordering of the demonstrations. In \Cref{tab:per_index},
we compute the GLM coefficients for  $S_i^R = s(R_q, R_i)$ when the performance is the response variable. 
For random sampling, we observe that this coefficient does not depend much on the position, while for RICES the coefficient increases from 0.01 (first rank) to 0.30 (4th rank) in captioning (and similarly in VQA, but to a lesser extent). As we saw earlier, this might be explained by the fact that this coefficient increases with more similar demonstrations. Another possibility is that M-ICL relies more on later ranks. The lines "RICE reverse" show that the latter explanation is truer, since by reversing the RICES order the coefficient still increases (to some extent) with the rank of the demonstration.

\begin{table}[t]
\small
\centering
\begin{tabular}{ll|rrrr}
\toprule
\multirow{2}{*}{Dataset} & \multirow{2}{*}{Sampling} & \multicolumn{4}{c}{$S^R_i \sim$ perf} \\
& & $S^R_1$ & $S^R_2$ & $S^R_3$ & $S^R_4$ \\
\midrule
\multirow{3}{*}{COCO} & Random & 0.26 & 0.26 & 0.25 & 0.18 \\
& RICES & 0.01 & 0.06 & 0.14 & 0.30 \\
& RICES Reverse & 0.11 & 0.13 & 0.09 & 0.20 \\
\midrule
\multirow{3}{*}{VQA} & Random & 0.10 & 0.13 & 0.22 & 0.21 \\
& RICES & 0.10 & 0.15 & 0.16 & 0.20 \\
& RICES Reverse & 0.15 & 0.21 & 0.19 & 0.06 \\
\bottomrule
\end{tabular}
\caption{\textbf{Influence of demonstrations on the performance based on their position.}
GLM coefficients (with the score as the response variable) of the similarity of each demonstration following his position. All coefficients have a p-value $<$ 0.01
}
\label{tab:per_index}
\end{table}

\begin{figure}[t]
\vspace{-1.5em}
    \centering
    \includegraphics[width=\linewidth]{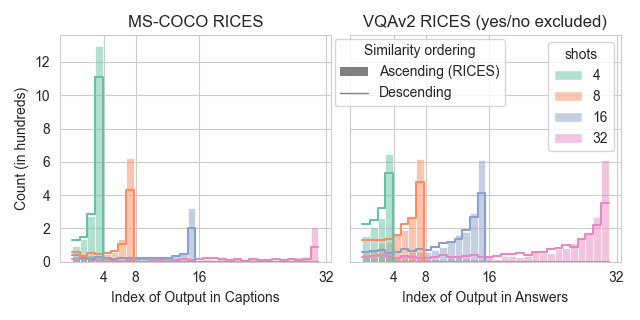}
    \vspace{-2em}
    \caption{\textbf{RICES M-ICL tends to copy the output of recent demonstrations} Count for RICES M-ICL of exact match of output with one of the demonstrations responses, out of 5000 analyzed items. As a patch, we have RICES in classic setup, and as a line, the same demonstrations are ordered by most similar to least.}
    \vspace{-0.8em}
    \label{fig:repetition}
\end{figure}

To further analyze the impact of this recency phenomenon, we compare the outputs of the model against each demonstration's output. Where there is a complete match between an entire demonstration's response and the full output produced by the model
For multiple matches, only the last one is recorded. Yes/no answers are excluded since in their frequency in VQA would skew the results.
This allows us to measure the extent with which a demonstration response is replicated in the model output. Although we observed that exact copies are extremely rare for random sampling (not shown here), the RICES method shows a frequent replication of the last demonstration (as depicted by the bars in \Cref{fig:repetition}). For VQA, the final context response is used 12\% of the cases, regardless of the number of shots. For captioning, this ranges from 24\% with four shots to 4\% with 32 shots. We compare with a variation of RICES where demonstrations are arranged from most to least similar (depicted by the lines). In this setup, the model less frequently replicates the last output, yet the same trend remains, indicating that the model tends to replicate the outputs of the more recent demonstrations over the more similar ones. This demonstrates that when M-ICL is faced with similar demonstrations, a recency bias leans towards replicating the output of the latest ones rather than the most similar. 

\section{Conclusion}
We propose a framework to study ICL in a multimodal context. Our study reveals that M-ICL is primarily text-driven, and that images in the context have little impact on the overall performance. This is exacerbated when using RICES to improve M-ICL. 
We also show that the reason of the success of similarity-based M-ICL is partially due to the fact that such techniques retrieve responses which are more similar to the target one rather than merely retrieving more related demonstrations. The practical consequences are that for classification-based tasks, M-ICL is useless when using RICES, and that for open-ended generation, there is still a gap that could be leveraged between RICES-retrieved responses and ideal ones.
In addition, we show that M-ICL suffers from different biases, such as the ability to replicate the last example in the demonstrations.
Our work sheds light on several limitations and suggests that there is room for improvement regarding M-ICL. Current M-ICL improvements can be brought by M-ICL variants or prompting strategies \cite{shukor_beyond_2023,mitra2023compositional,yu_coca_2022}, or better training datasets \cite{zhao2024mmicl,laurencon_obelics_2023}. Our work suggests that working on better retrieval and reducing the biases 
(\eg recency) 
would also benefit this line of models.
Finally, while our findings hold for the best open-source M-ICL models, we recognize it would be  important to study more powerful models such as GPT4-V \cite{gpt4} and Gemini \cite{gemini} to check if our conclusion still hold.

\section{Acknowledgments}
This work was partly funded by the ANR-23-PEIA-0008, PEPR IA, project "Principes théoriques et algorithmiques de l’apprentissage frugal (SHARP)," and received computing AI and storage resources from GENCI at IDRIS on the Jean Zay supercomputer's V100/A100 partition through grant 2023-AD011014764.

\newpage
{
    \small
    \bibliographystyle{ieeenat_fullname}
    \bibliography{main}
}

\section{Appendix}

\subsection{Consideration on different behaviour of IDEFICS and OpenFlamingo}\label{par:openflamingo}
The two open-source models, IDEFICS~\cite{laurencon_obelics_2023} and OpenFlamingo~\cite{awadalla_openflamingo_2023}, are both implementations of the model proposed by~\cite{alayrac_flamingo_2022}. Despite sharing the same architecture, our analysis, as observable in ~\ref{tab:openflamingo} and ~\ref{tab:remove_modality_full} , reveals distinct behaviors between the two models when subjected to image removal or random image swapping.
OpenFlamingo demonstrates a slight decrease in performance when removing or swapping images compared to the godlen prompt, indicating minimal impact from perturbations and recognising task, but not focusing on the image-text mapping. On the other hand, IDEFICS exhibits a larger performance drop without images and with random images experiences even further degradation with an increase in the number of shots.

\begin{table}[h]
    \footnotesize
    \centering
    \begin{tabular}{llrrrr}
\toprule
 & num shots & 4 & 8 & 16 & 32 \\
dataset & Prompt &  &  &  &  \\
\midrule
\multirow[t]{3}{*}{Flickr30k} & W/o image & 61.11 & 63.45 & 62.57 & 61.66 \\
 & Rnd. image & 51.04 & 53.15 & 58.20 & 59.07 \\
 & Base & 60.92 & 62.42 & 64.05 & 63.03 \\
\cline{1-6}
\multirow[t]{3}{*}{ImageNet 1k} & W/o image & 25.67 & 24.05 & 20.93 & 16.43 \\
 & Rnd. image & 11.09 & 7.73 & 6.16 & 5.18 \\
 & Base & 22.55 & 21.54 & 18.73 & 16.11 \\
\cline{1-6}
\multirow[t]{3}{*}{MS-COCO} & W/o image & 83.33 & 88.68 & 93.36 & 94.50 \\
 & Rnd. image & 76.31 & 84.89 & 90.55 & NaN \\
 & Base & 84.43 & 91.34 & 96.52 & NaN \\
\cline{1-6}
\multirow[t]{3}{*}{rendered SST-2} & W/o image & 10.70 & 29.87 & 11.19 & 14.22 \\
 & Rnd. image & 53.48 & 61.29 & 60.63 & 56.79 \\
 & Base & 53.44 & 59.94 & 60.53 & 57.91 \\
\cline{1-6}
\bottomrule
\end{tabular}

    \caption{
    Evaluation results using OpenFlamingo 9B and  demonstrations sampled uniformly at random across four image-to-text datasets using 0, 4, 8, 16, and 32 in-context demonstrations.
Depicted various prompt modifications, such as removing one modality (either the image or the question) or replacing it with a different random instance from the training dataset.
    }\label{tab:openflamingo}
\end{table}

The disparity in behavior between the two models can likely be attributed to differences in their training datasets. IDEFICS was trained on the OBELICS~\cite{laurencon_obelics_2023} dataset, which contains longer, more contextual texts and extracts data directly from the HTML DOM tree, thus providing cleaner data free from ads and spam. This method ensures higher document quality, comparable to renowned datasets like The Pile and Wikipedia. Furthermore, OBELICS addresses the issue of image duplication present in Multimodal C4, in which only 60\% of images are unique, thus offering a higher quality and more efficient training dataset. In contrast, OpenFlamingo was trained on the shorter, less detailed texts of Multimodal C4.

Given that IDEFICS generally achieves better scores and is more responsive to ICL, we have chosen to focus our study on this model.

\paragraph{Comparaison with~\citet{chen_understanding_2023}}
The findings presented by~\citet{chen_understanding_2023}, corroborate the behavioral differences between the two models that we observed. However, their study emphasizes the behavior of OpenFlamingo and concludes that ICLis primarily driven by text, as it appears insensitive to changes in images. Our observations regarding VQA align with this: ICL indeed seems to be driven predominantly by text. However, we note a different pattern in image-to-text tasks, where ICL does respond to visual elements. Nonetheless, when text is also available, it tends to become the dominant factor influencing the model's responses.

\subsection{Balanced sampling}
In ~\Cref{sec:knn}, we demonstrated that the performance of RICES ICL improves significantly due to a majority voting process that selects the most common label in a given context. To better understand how label imbalance impacts this, we conducted experiments in a binary classification framework, adjusting the sampling method to ensure an equal number of demonstrations from each class in the context. For random sampling, the demonstrations were arranged without specific order, while for RICES, we selected the closest demonstrations from each class and sorted them by increasing similarity. In~\cref{tab:balanced_sampling}, we found the following order of performance from worst to best: random sampling (comparaison point), balanced random sampling (+1.74\% improvement), balanced RICES sampling (+8.40\% improvement), and RICES sampling alone (+18.90\% improvement). This suggests that while balancing the samples improves performance in random contexts, the balanced RICES approach yields only half the performance boost compared to using RICES alone. Therefore, we can infer that while example similarity contributes to model performance, the distribution of labels plays an important role.

\begin{table}[h]
    \centering
    \footnotesize
    \begin{tabular}{llrrrr}
\toprule
 & num shots & 4 & 8 & 16 & 32 \\
dataset & sampling &  &  &  &  \\
\midrule
\multirow[t]{4}{*}{Hateful Memes} & RICES & 60.50 & 62.30 & 63.40 & 62.60 \\
 & Balanced rnd. & 53.30 & 53.37 & 55.03 & 55.17 \\
 & Balanced RICES & 54.60 & 56.10 & 58.30 & 57.70 \\
 & Random & 50.57 & 50.93 & 52.00 & 53.77 \\
\cline{1-6}
\multirow[t]{4}{*}{rendered SST-2} & RICES & 75.80 & 84.14 & 82.84 & 80.18 \\
 & Balanced rnd. & 57.07 & 57.27 & 58.11 & 58.85 \\
 & Balanced RICES & 61.46 & 70.74 & 77.30 & 80.34 \\
 & Random & 56.41 & 56.81 & 57.62 & 58.67 \\
\cline{1-6}
\bottomrule
\end{tabular}

    \caption{Evaluation results using IDEFICS 9B across two binary classification vision-language datasets using 0, 4, 8, 16, and 32 in-context demonstrations.
Depicted various sampling methods, random sampling (Random), RICES and their balanced counterparts.}\label{tab:balanced_sampling}
\end{table}

\begin{figure*}[ht]
    \centering
    \includegraphics[width=\linewidth]{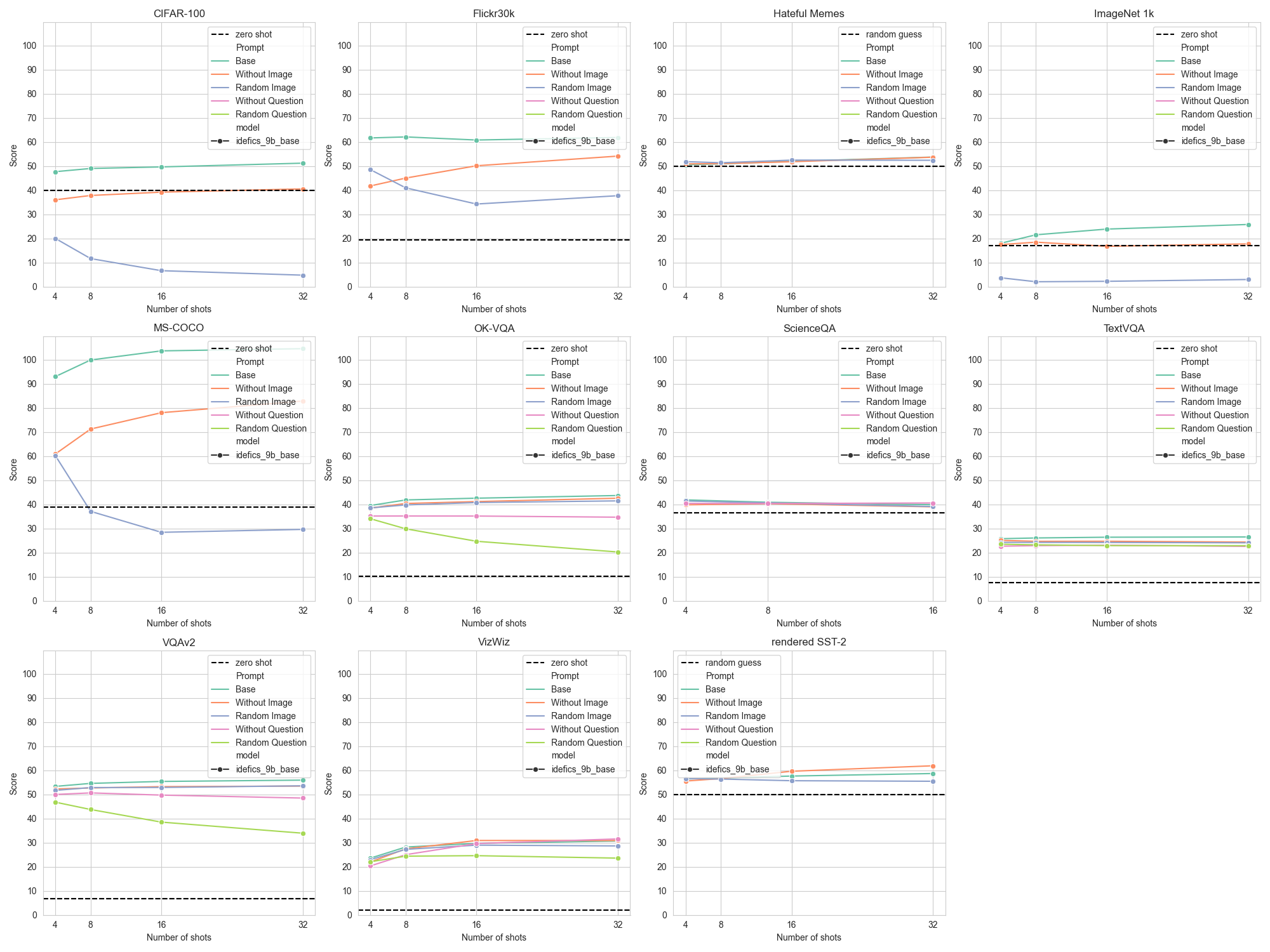}
    \caption{Full evaluation results using IDEFICS 9B and  demonstrations sampled uniformly at random across twelve
vision-language datasets using 0, 4, 8, 16, and 32 in-context demonstrations.
Depicted various prompt modifications, such as removing one modality (either the image or the question) or replacing it with a different random instance from the training dataset.}
    \label{fig:remove_modality}
\end{figure*}
\begin{figure*}[ht]
    \centering
    \includegraphics[width=\linewidth]{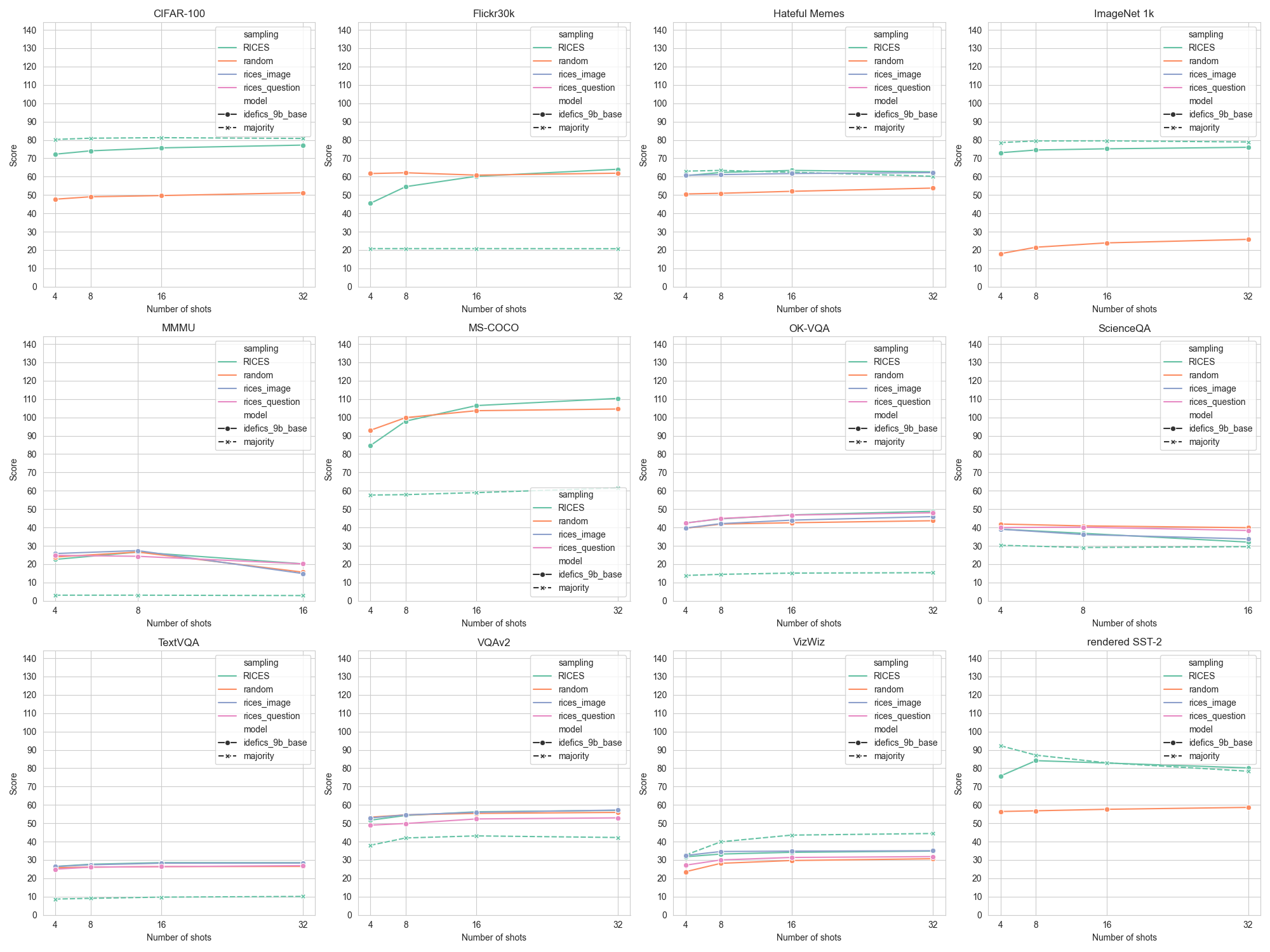}
    \caption{Full evaluation results using IDEFICS 9B and base prompt across twelve
vision-language datasets using 0, 4, 8, 16, and 32 in-context demonstrations. Depicted the scores of random sampling (Random) and RICES in is standard form or using only one modality for similarity function (rices\_modality)}\label{fig:rices_key_full}
\end{figure*}
\begin{figure*}[ht]
    \centering
    \includegraphics[width=\linewidth]{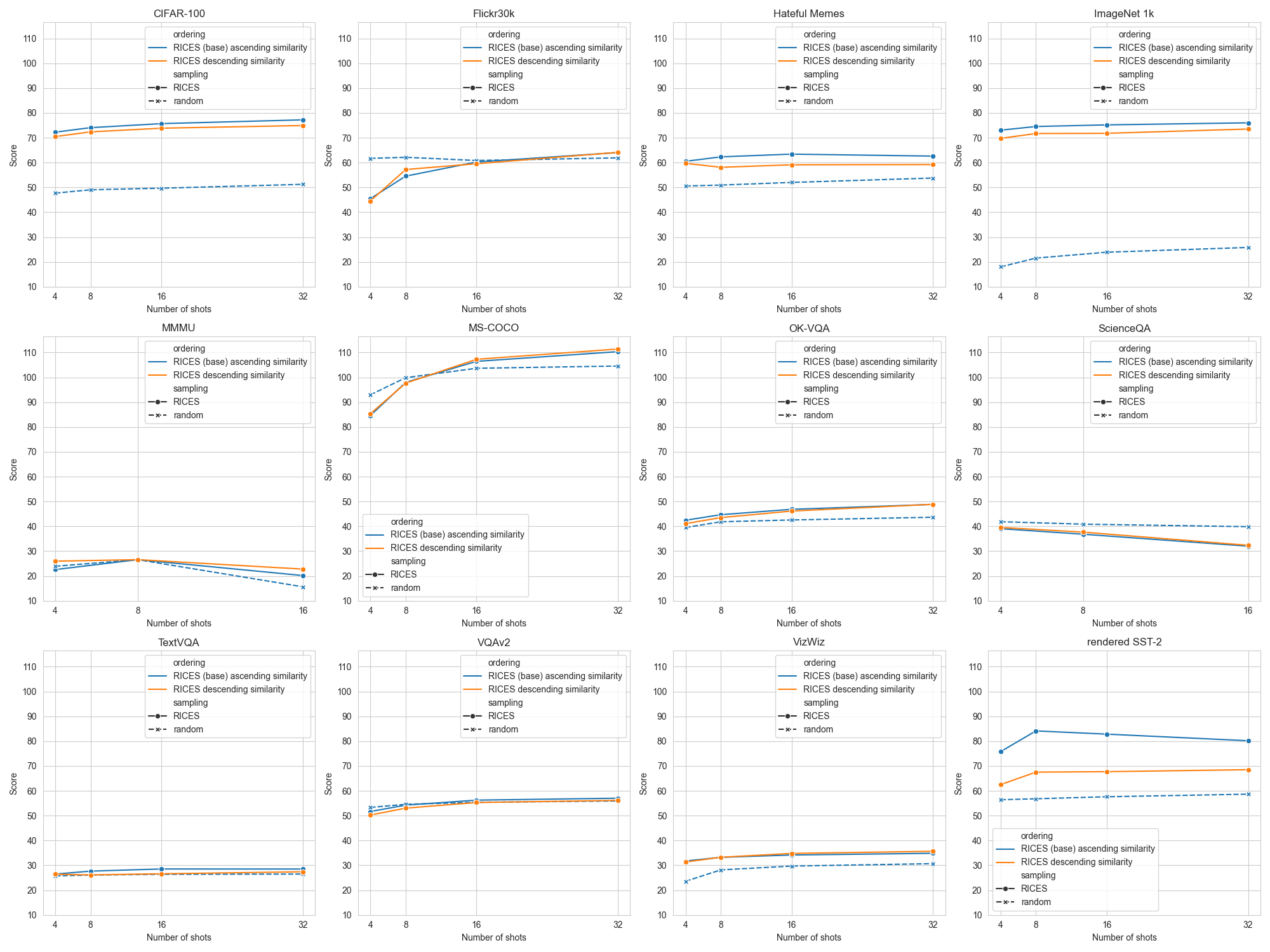}
    \caption{Evaluation results using IDEFICS 9B and base prompt across twelve
vision-language datasets using 0, 4, 8, 16, and 32 in-context demonstrations. Comparison of RICES with default order of demonstration (ascending) and a variant with descending similarity ordering.}\label{fig:enter-label}
\end{figure*}
\begin{figure*}[ht]
    \centering
    \includegraphics[width=\linewidth]{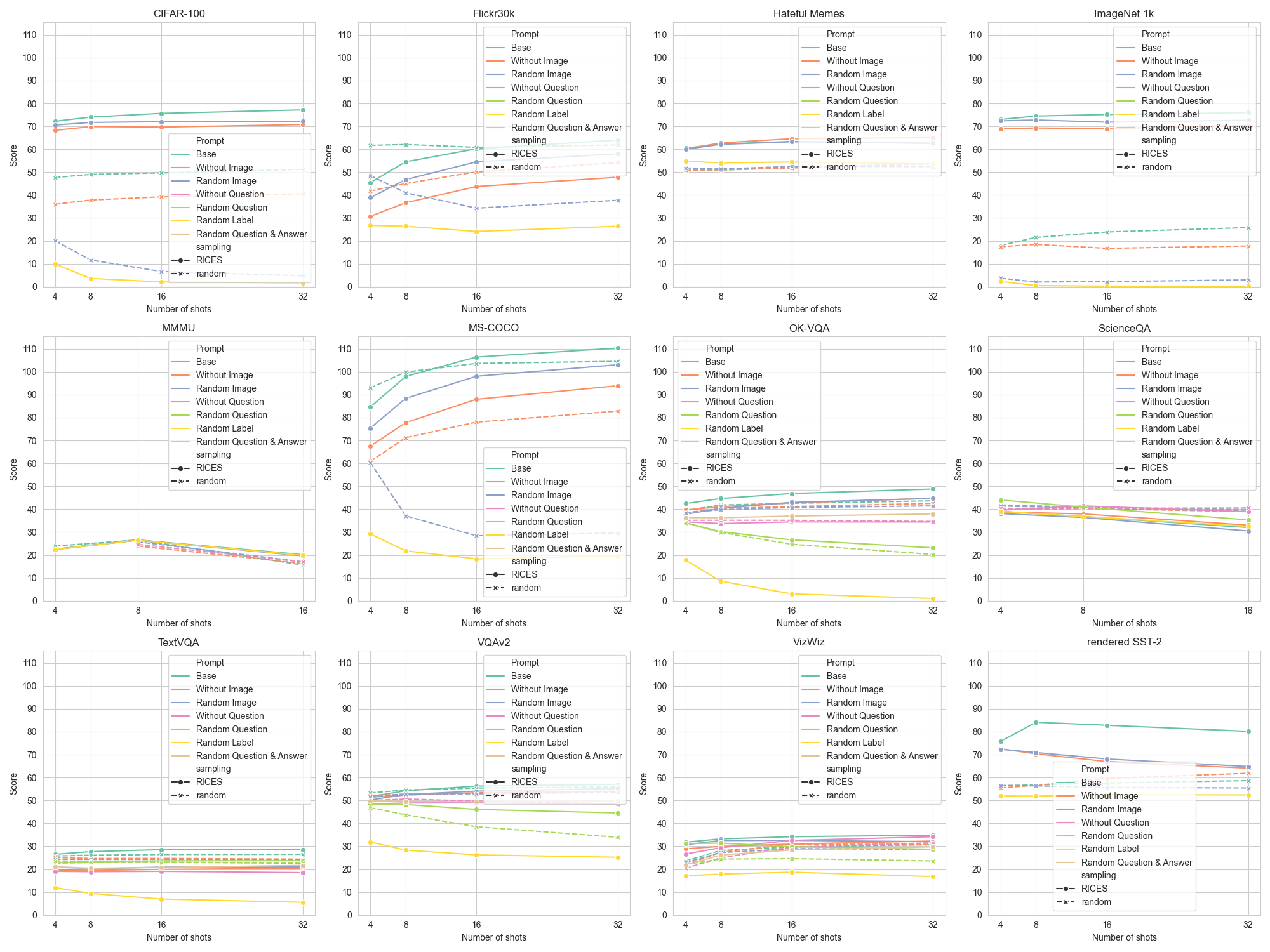}
    \caption{ull evaluation results using IDEFICS 9B and demonstrations sampled with RICES across twelve
vision-language datasets using 0, 4, 8, 16, and 32 in-context demonstrations.
Depicted various prompt modifications, such as removing one modality (either the image or the question) or replacing it with a different random instance from the training dataset.}
    \label{fig:rices_no_modality_full}
\end{figure*}

\begin{table}[ht]
    \centering
    \footnotesize
    \begin{tabular}{llrrrr}
\toprule
 & Shots & 4 & 8 & 16 & 32 \\
Dataset & Sampling &  &  &  &  \\
\midrule
\multirow[t]{2}{*}{CIFAR-100} & R. image & 72.24 & 74.06 & 75.68 & 77.20 \\
 & Random & 47.70 & 49.03 & 49.68 & 51.23 \\
\cline{1-6}
\multirow[t]{2}{*}{Flickr30k} & RICES & 45.48 & 54.54 & 60.21 & 64.03 \\
 & Random & 61.69 & 62.12 & 60.83 & 61.89 \\
\cline{1-6}
\multirow[t]{4}{*}{Hateful Memes} & RICES & 60.50 & 62.30 & 63.40 & 62.60 \\
 & Random & 50.57 & 50.93 & 52.00 & 53.77 \\
 & R. image & 60.80 & 61.10 & 61.70 & 62.20 \\
 & R. OCR & 59.80 & 61.70 & 62.30 & 62.80 \\
\cline{1-6}
\multirow[t]{2}{*}{ImageNet 1k} & RICES & 73.04 & 74.52 & 75.18 & 76.00 \\
 & Random & 18.01 & 21.53 & 23.90 & 25.81 \\
\cline{1-6}
\multirow[t]{4}{*}{MMMU} & RICES & 22.60 & 26.60 & 20.20 & NaN \\
 & Random & 23.93 & 26.60 & 15.67 & NaN \\
 & R. image & 25.80 & 27.40 & 14.90 & NaN \\
 & R. question & 24.80 & 24.30 & 20.20 & NaN \\
\cline{1-6}
\multirow[t]{2}{*}{MS-COCO} & RICES & 84.65 & 97.98 & 106.44 & 110.36 \\
 & Random & 92.98 & 99.88 & 103.66 & 104.57 \\
\cline{1-6}
\multirow[t]{4}{*}{OK-VQA} & RICES & 42.47 & 44.70 & 46.87 & 48.84 \\
 & Random & 39.54 & 41.85 & 42.58 & 43.68 \\
 & R. image & 39.71 & 42.10 & 44.00 & 45.94 \\
 & R. question & 42.35 & 44.92 & 46.74 & 48.02 \\
\cline{1-6}
\multirow[t]{4}{*}{ScienceQA} & RICES & 39.07 & 36.84 & 32.03 & NaN \\
 & Random & 41.88 & 40.89 & 39.88 & NaN \\
 & R. image & 39.07 & 36.14 & 33.76 & NaN \\
 & R. question & 39.96 & 40.16 & 38.37 & NaN \\
\cline{1-6}
\multirow[t]{4}{*}{TextVQA} & RICES & 26.48 & 27.67 & 28.54 & 28.51 \\
 & Random & 25.77 & 26.09 & 26.40 & 26.50 \\
 & R. image & 26.39 & 27.38 & 28.24 & 28.33 \\
 & R. question & 24.97 & 26.10 & 26.37 & 26.91 \\
\cline{1-6}
\multirow[t]{4}{*}{VQAv2} & RICES & 51.68 & 54.25 & 56.26 & 57.04 \\
 & Random & 53.33 & 54.58 & 55.39 & 55.93 \\
 & R. image & 52.92 & 54.57 & 55.98 & 57.20 \\
 & R. question & 48.99 & 49.88 & 52.37 & 52.95 \\
\cline{1-6}
\multirow[t]{4}{*}{VizWiz} & RICES & 31.75 & 33.23 & 34.17 & 34.85 \\
 & Random & 23.58 & 28.18 & 29.71 & 30.69 \\
 & R. image & 32.45 & 34.61 & 34.82 & 35.03 \\
 & R. question & 27.15 & 30.02 & 31.37 & 31.83 \\
\cline{1-6}
\multirow[t]{2}{*}{rendered SST-2} & RICES & 75.80 & 84.14 & 82.84 & 80.18 \\
 & Random & 56.41 & 56.81 & 57.62 & 58.67 \\
\cline{1-6}
\bottomrule
\end{tabular}

    \caption{Full evaluation results using IDEFICS 9B and base prompt across twelve
vision-language datasets using 0, 4, 8, 16, and 32 in-context demonstrations. Depicted the scores of random sampling (Random) and RICES in is standard form or using only one modality for similarity function (R. modality)}\label{tab:rices_key_full}
\end{table}
\begin{table}[ht]
    \centering
    \footnotesize
    \begin{tabular}{llrrrr}
\toprule
 & Shots & 4 & 8 & 16 & 32 \\
Dataset & Prompt &  &  &  &  \\
\midrule
\multirow[t]{3}{*}{CIFAR-100} & W/o image & 36.03 & 37.84 & 39.21 & 40.57 \\
 & Rnd. image & 20.05 & 11.67 & 6.65 & 4.79 \\
 & Base & 47.70 & 49.03 & 49.68 & 51.23 \\
\cline{1-6}
\multirow[t]{3}{*}{Flickr30k} & W/o image & 41.78 & 45.05 & 50.15 & 54.16 \\
 & Rnd. image & 48.54 & 40.99 & 34.29 & 37.75 \\
 & Base & 61.69 & 62.12 & 60.83 & 61.89 \\
\cline{1-6}
\multirow[t]{3}{*}{Hateful Memes} & W/o image & 50.93 & 51.03 & 51.83 & 53.70 \\
 & Rnd. image & 51.87 & 51.40 & 52.50 & 52.43 \\
 & Base & 50.57 & 50.93 & 52.00 & 53.77 \\
\cline{1-6}
\multirow[t]{3}{*}{ImageNet 1k} & W/o image & 17.46 & 18.47 & 16.77 & 17.77 \\
 & Rnd. image & 3.72 & 2.07 & 2.25 & 2.99 \\
 & Base & 18.01 & 21.53 & 23.90 & 25.81 \\
\cline{1-6}
\multirow[t]{3}{*}{MS-COCO} & W/o image & 60.87 & 71.25 & 78.02 & 82.86 \\
 & Rnd. image & 60.29 & 37.12 & 28.43 & 29.63 \\
 & Base & 92.98 & 99.88 & 103.66 & 104.57 \\
\cline{1-6}
\multirow[t]{5}{*}{OK-VQA} & W/o image & 38.48 & 40.39 & 41.17 & 42.56 \\
 & W/o question & 35.18 & 35.19 & 35.17 & 34.70 \\
 & Rnd. image & 38.54 & 39.78 & 40.77 & 41.46 \\
 & Rnd. question & 34.06 & 29.88 & 24.72 & 20.27 \\
 & Base & 39.54 & 41.85 & 42.58 & 43.68 \\
\cline{1-6}
\multirow[t]{4}{*}{ScienceQA} & W/o image & 39.83 & 40.31 & 38.99 & NaN \\
 & W/o question & 40.41 & 40.37 & 40.59 & NaN \\
 & Rnd. image & 41.41 & 40.64 & 39.12 & NaN \\
 & Base & 41.88 & 40.89 & 39.88 & NaN \\
\cline{1-6}
\multirow[t]{5}{*}{TextVQA} & W/o image & 25.08 & 24.69 & 24.71 & 24.38 \\
 & W/o question & 22.66 & 22.90 & 23.08 & 22.58 \\
 & Rnd. image & 24.25 & 24.26 & 24.22 & 24.08 \\
 & Rnd. question & 23.49 & 23.23 & 22.92 & 22.87 \\
 & Base & 25.77 & 26.09 & 26.40 & 26.50 \\
\cline{1-6}
\multirow[t]{5}{*}{VQAv2} & W/o image & 52.26 & 52.67 & 53.22 & 53.47 \\
 & W/o question & 49.98 & 50.63 & 49.73 & 48.49 \\
 & Rnd. image & 51.67 & 52.84 & 52.90 & 53.57 \\
 & Rnd. question & 46.80 & 43.75 & 38.52 & 33.92 \\
 & Base & 53.33 & 54.58 & 55.39 & 55.93 \\
\cline{1-6}
\multirow[t]{5}{*}{VizWiz} & W/o image & 21.96 & 27.44 & 30.90 & 30.89 \\
 & W/o question & 20.36 & 25.02 & 29.63 & 31.55 \\
 & Rnd. image & 22.94 & 27.22 & 28.97 & 28.65 \\
 & Rnd. question & 22.08 & 24.40 & 24.60 & 23.59 \\
 & Base & 23.58 & 28.18 & 29.71 & 30.69 \\
\cline{1-6}
\multirow[t]{3}{*}{rendered SST-2} & W/o image & 55.55 & 56.57 & 59.61 & 61.88 \\
 & Rnd. image & 56.57 & 56.37 & 55.69 & 55.46 \\
 & Base & 56.41 & 56.81 & 57.62 & 58.67 \\
\cline{1-6}
\bottomrule
\end{tabular}

    \caption{Full evaluation results using IDEFICS 9B and  demonstrations sampled uniformly at random across twelve
vision-language datasets using 0, 4, 8, 16, and 32 in-context demonstrations.
Depicted various prompt modifications, such as removing one modality (either the image or the question) or replacing it with a different random instance from the training dataset.}\label{tab:remove_modality_full}
\end{table}
\begin{table}[ht]
    \centering
    \footnotesize
    \begin{tabular}{llrrrr}
\toprule
 & Shots & 4 & 8 & 16 & 32 \\
Dataset & ordering &  &  &  &  \\
\midrule
\multirow[t]{2}{*}{CIFAR-100} & ascending & 72.24 & 74.06 & 75.68 & 77.20 \\
 & descending & 70.46 & 72.36 & 73.84 & 74.92 \\
\cline{1-6}
\multirow[t]{2}{*}{Flickr30k} & ascending & 45.48 & 54.54 & 60.21 & 64.03 \\
 & descending & 44.56 & 57.21 & 59.53 & 64.11 \\
\cline{1-6}
\multirow[t]{2}{*}{Hateful Memes} & ascending & 60.50 & 62.30 & 63.40 & 62.60 \\
 & descending & 59.70 & 58.10 & 59.10 & 59.20 \\
\cline{1-6}
\multirow[t]{2}{*}{ImageNet 1k} & ascending & 73.04 & 74.52 & 75.18 & 76.00 \\
 & descending & 69.74 & 71.70 & 71.78 & 73.50 \\
\cline{1-6}
\multirow[t]{2}{*}{MMMU} & ascending & 22.60 & 26.60 & 20.20 & NaN \\
 & descending & 26.00 & 26.60 & 22.80 & NaN \\
\cline{1-6}
\multirow[t]{2}{*}{MS-COCO} & ascending & 84.65 & 97.98 & 106.44 & 110.36 \\
 & descending & 85.37 & 97.68 & 107.28 & 111.41 \\
\cline{1-6}
\multirow[t]{2}{*}{OK-VQA} & ascending & 42.47 & 44.70 & 46.87 & 48.84 \\
 & descending & 41.09 & 43.54 & 46.16 & 48.88 \\
\cline{1-6}
\multirow[t]{2}{*}{ScienceQA} & ascending & 39.07 & 36.84 & 32.03 & NaN \\
 & descending & 39.56 & 37.68 & 32.37 & NaN \\
\cline{1-6}
\multirow[t]{2}{*}{TextVQA} & ascending & 26.48 & 27.67 & 28.54 & 28.51 \\
 & descending & 26.42 & 26.14 & 26.61 & 27.40 \\
\cline{1-6}
\multirow[t]{2}{*}{VQAv2} & ascending & 51.68 & 54.25 & 56.26 & 57.04 \\
 & descending & 50.26 & 53.04 & 55.30 & 56.16 \\
\cline{1-6}
\multirow[t]{2}{*}{VizWiz} & ascending & 31.75 & 33.23 & 34.17 & 34.85 \\
 & descending & 31.33 & 33.26 & 34.80 & 35.66 \\
\cline{1-6}
\multirow[t]{2}{*}{rendered SST-2} & ascending & 75.80 & 84.14 & 82.84 & 80.18 \\
 & descending & 62.52 & 67.54 & 67.72 & 68.52 \\
\cline{1-6}
\bottomrule
\end{tabular}

    \caption{
Evaluation results using IDEFICS 9B and base prompt across twelve
vision-language datasets using 0, 4, 8, 16, and 32 in-context demonstrations. Comparison of RICES with default order of demonstration (ascending) and a variant with descending similarity ordering.}\label{tab:rices_reverse}
\end{table}

\begin{table}[ht]
    \centering
    \footnotesize
    \begin{tabular}{llrrrr}
\toprule
 & Shots & 4 & 8 & 16 & 32 \\
Dataset & Variant &  &  &  &  \\
\midrule
\multirow[t]{3}{*}{CIFAR-100} &   Rnd. S. LMM & 47.70 & 49.03 & 49.68 & 51.23 \\
 &  RICES LMM & 72.24 & 74.06 & 75.68 & 77.20 \\
 & RICES KNN & 80.28 & 80.96 & 81.24 & 80.82 \\
\cline{1-6}
\multirow[t]{3}{*}{Flickr30k} &   Rnd. S. LMM & 61.69 & 62.12 & 60.83 & 61.89 \\
 &  RICES LMM & 45.48 & 54.54 & 60.21 & 64.03 \\
 & RICES KNN & 20.77 & 20.77 & 20.77 & 20.73 \\
\cline{1-6}
\multirow[t]{3}{*}{Hateful Memes} &   Rnd. S. LMM & 50.57 & 50.93 & 52.00 & 53.77 \\
 &  RICES LMM & 60.50 & 62.30 & 63.40 & 62.60 \\
 & RICES KNN & 63.00 & 63.40 & 62.40 & 60.20 \\
\cline{1-6}
\multirow[t]{3}{*}{ImageNet 1k} &   Rnd. S. LMM & 18.01 & 21.53 & 23.90 & 25.81 \\
 &  RICES LMM & 73.04 & 74.52 & 75.18 & 76.00 \\
 & RICES KNN & 78.58 & 79.46 & 79.52 & 78.90 \\
\cline{1-6}
\multirow[t]{3}{*}{MMMU} &   Rnd. S. LMM & 23.93 & 26.60 & 15.67 & NaN \\
 &  RICES LMM & 22.60 & 26.60 & 20.20 & NaN \\
 & RICES KNN & 3.10 & 3.10 & 2.90 & NaN \\
\cline{1-6}
\multirow[t]{3}{*}{MS-COCO} &   Rnd. S. LMM & 92.98 & 99.88 & 103.66 & 104.57 \\
 &  RICES LMM & 84.65 & 97.98 & 106.44 & 110.36 \\
 & RICES KNN & 57.69 & 57.90 & 59.00 & 61.55 \\
\cline{1-6}
\multirow[t]{3}{*}{OK-VQA} &   Rnd. S. LMM & 39.54 & 41.85 & 42.58 & 43.68 \\
 &  RICES LMM & 42.47 & 44.70 & 46.87 & 48.84 \\
 & RICES KNN & 13.86 & 14.46 & 15.14 & 15.35 \\
\cline{1-6}
\multirow[t]{3}{*}{ScienceQA} &   Rnd. S. LMM & 41.88 & 40.89 & 39.88 & NaN \\
 &  RICES LMM & 39.07 & 36.84 & 32.03 & NaN \\
 & RICES KNN & 30.29 & 29.10 & 29.55 & NaN \\
\cline{1-6}
\multirow[t]{3}{*}{TextVQA} &   Rnd. S. LMM & 25.77 & 26.09 & 26.40 & 26.50 \\
 &  RICES LMM & 26.48 & 27.67 & 28.54 & 28.51 \\
 & RICES KNN & 8.69 & 9.09 & 9.75 & 10.13 \\
\cline{1-6}
\multirow[t]{3}{*}{VQAv2} &   Rnd. S. LMM & 53.33 & 54.58 & 55.39 & 55.93 \\
 &  RICES LMM & 51.68 & 54.25 & 56.26 & 57.04 \\
 & RICES KNN & 38.01 & 42.01 & 43.12 & 42.25 \\
\cline{1-6}
\multirow[t]{3}{*}{VizWiz} &   Rnd. S. LMM & 23.58 & 28.18 & 29.71 & 30.69 \\
 &  RICES LMM & 31.75 & 33.23 & 34.17 & 34.85 \\
 & RICES KNN & 32.66 & 39.91 & 43.55 & 44.43 \\
\cline{1-6}
\multirow[t]{3}{*}{rendered SST-2} &   Rnd. S. LMM & 56.41 & 56.81 & 57.62 & 58.67 \\
 &  RICES LMM & 75.80 & 84.14 & 82.84 & 80.18 \\
 & RICES KNN & 92.26 & 87.12 & 82.96 & 78.38 \\
\cline{1-6}
\bottomrule
\end{tabular}

    \caption{Evaluation results using IDEFICS 9B across twelve
vision-language datasets using 0, 4, 8, 16, and 32 in-context demonstrations.
Depicted M-ICL with random sampling (Rnd. S. LMM), M-ICL with RICES sampling (RICES LMM) and the majority voting baseline (RICES KNN)}\label{tab:knn_table}
\end{table}
\begin{table}[ht]
    \centering
    \footnotesize
    \begin{tabular}{llrrrr}
\toprule
 & Shots & 4 & 8 & 16 & 32 \\
Dataset & Prompt &  &  &  &  \\
\midrule
\multirow[t]{4}{*}{CIFAR-100} & W/o image & 68.28 & 69.88 & 69.68 & 70.80 \\
 & Rnd. image & 70.59 & 71.71 & 72.07 & 72.18 \\
 & Rnd. label & 9.91 & 3.63 & 2.09 & 1.72 \\
 & Base & 72.24 & 74.06 & 75.68 & 77.20 \\
\cline{1-6}
\multirow[t]{4}{*}{Flickr30k} & W/o image & 30.75 & 36.66 & 43.75 & 47.83 \\
 & Rnd. image & 38.88 & 46.78 & 54.52 & 58.04 \\
 & Rnd. label & 26.80 & 26.40 & 24.12 & 26.42 \\
 & Base & 45.48 & 54.54 & 60.21 & 64.03 \\
\cline{1-6}
\multirow[t]{4}{*}{Hateful Memes} & W/o image & 60.10 & 62.80 & 64.60 & 65.10 \\
 & Rnd. image & 60.00 & 62.17 & 63.27 & 62.70 \\
 & Rnd. label & 54.77 & 54.10 & 54.43 & 53.67 \\
 & Base & 60.50 & 62.30 & 63.40 & 62.60 \\
\cline{1-6}
\multirow[t]{4}{*}{ImageNet 1k} & W/o image & 68.94 & 69.28 & 69.02 & 70.42 \\
 & Rnd. image & 72.41 & 72.79 & 71.84 & 72.67 \\
 & Rnd. label & 2.32 & 0.51 & 0.23 & 0.14 \\
 & Base & 73.04 & 74.52 & 75.18 & 76.00 \\
\cline{1-6}
\multirow[t]{4}{*}{MMMU} & W/o image & 22.40 & 26.40 & 19.90 & NaN \\
 & Rnd. image & 22.47 & 26.47 & 19.67 & NaN \\
 & Rnd. label & 22.47 & 26.47 & 19.90 & NaN \\
 & Base & 22.60 & 26.60 & 20.20 & NaN \\
\cline{1-6}
\multirow[t]{4}{*}{MS-COCO} & W/o image & 67.58 & 77.81 & 88.01 & 93.93 \\
 & Rnd. image & 75.42 & 88.40 & 98.06 & 103.08 \\
 & Rnd. label & 29.19 & 21.85 & 18.34 & 19.64 \\
 & Base & 84.65 & 97.98 & 106.44 & 110.36 \\
\cline{1-6}
\multirow[t]{6}{*}{OK-VQA} & W/o image & 39.69 & 41.11 & 42.76 & 44.82 \\
 & W/o quest. & 34.50 & 33.77 & 34.47 & 34.44 \\
 & Rnd. image & 37.83 & 40.37 & 43.01 & 44.72 \\
 & Rnd. label & 17.80 & 8.60 & 3.06 & 1.02 \\
 & Rnd. quest. & 34.11 & 30.20 & 26.66 & 23.18 \\
 & Base & 42.47 & 44.70 & 46.87 & 48.84 \\
\cline{1-6}
\multirow[t]{6}{*}{ScienceQA} & W/o image & 38.7 & 37.98 & 33.07 & NaN \\
 & W/o quest. & 39.56 & 41.45 & 38.92 & NaN \\
 & Rnd. image & 38.13 & 36.42 & 30.52 & NaN \\
 & Rnd. label & 39.07 & 36.84 & 32.52 & NaN \\
 & Rnd. quest. & 44.04 & 40.92 & 35.35 & NaN \\
 & Base & 39.07 & 36.84 & 32.03 & NaN \\
\cline{1-6}
\multirow[t]{6}{*}{TextVQA} & W/o image & 19.68 & 19.43 & 19.94 & 20.36 \\
 & W/o quest. & 19.05 & 18.90 & 19.04 & 18.51 \\
 & Rnd. image & 19.77 & 20.17 & 20.91 & 20.97 \\
 & Rnd. label & 11.97 & 9.44 & 6.96 & 5.56 \\
 & Rnd. quest. & 22.82 & 23.17 & 23.63 & 23.71 \\
 & Base & 26.48 & 27.67 & 28.54 & 28.51 \\
\cline{1-6}
\multirow[t]{6}{*}{VQAv2} & W/o image & 51.43 & 52.32 & 53.39 & 54.07 \\
 & W/o quest. & 48.47 & 48.90 & 48.96 & 48.38 \\
 & Rnd. image & 50.21 & 52.63 & 54.08 & 55.22 \\
 & Rnd. label & 31.87 & 28.32 & 26.24 & 25.20 \\
 & Rnd. quest. & 48.42 & 48.24 & 46.10 & 44.54 \\
 & Base & 51.68 & 54.25 & 56.26 & 57.04 \\
\cline{1-6}
\multirow[t]{6}{*}{VizWiz} & W/o image & 28.89 & 29.95 & 30.98 & 32.29 \\
 & W/o quest. & 26.62 & 29.39 & 32.54 & 34.17 \\
 & Rnd. image & 30.67 & 32.51 & 32.56 & 31.98 \\
 & Rnd. label & 17.11 & 17.86 & 18.70 & 16.78 \\
 & Rnd. quest. & 31.21 & 31.37 & 29.81 & 28.69 \\
 & Base & 31.75 & 33.23 & 34.17 & 34.85 \\
\cline{1-6}
\multirow[t]{4}{*}{rendered SST-2} & W/o image & 72.56 & 70.36 & 67.00 & 64.10 \\
 & Rnd. image & 72.35 & 70.93 & 68.11 & 64.83 \\
 & Rnd. label & 51.97 & 51.89 & 52.53 & 52.41 \\
 & Base & 75.80 & 84.14 & 82.84 & 80.18 \\
\cline{1-6}
\bottomrule
\end{tabular}

    \caption{Full evaluation results using IDEFICS 9B and demonstrations sampled with RICES across twelve
vision-language datasets using 0, 4, 8, 16, and 32 in-context demonstrations.
Depicted various prompt modifications, such as removing one modality (either the image or the question) or replacing it with a different random instance from the training dataset.}\label{tab:rices_no_modality}
\end{table}

\begin{table}[ht]
    \centering
    \footnotesize
    \begin{tabular}{lr}
\toprule
Dataset & Zero-shot score \\
\midrule
ScienceQA & 36.39 \\
MMMU & 4.37 \\
MS-COCO & 38.94 \\
Flickr30k & 19.44 \\
OK-VQA & 10.29 \\
VQAv2 & 6.66 \\
VizWiz & 2.16 \\
ImageNet 1k & 16.98 \\
Hateful Memes & 0.00 \\
TextVQA & 7.66 \\
rendered SST-2 & 0.02 \\
CIFAR-100 & 39.98 \\
\bottomrule
\end{tabular}
    \caption{Full evaluation results using IDEFICS 9B across twelve
vision-language datasets using no demonstrations.}\label{tab:zero_shot}
\end{table}

\begin{table}[ht]
    \centering
    \footnotesize
    \begin{tabular}{lr}
\toprule
Dataset & Oracle RICES score \\
\midrule
CIFAR-100 & 91.98 \\
Flickr30k & 76.53 \\
Hateful Memes & 100.00 \\
ImageNet 1k & 99.56 \\
MMMU & 19.30 \\
MS-COCO & 139.03 \\
OK-VQA & 75.90 \\
ScienceQA & 35.05 \\
TextVQA & 49.79 \\
VQAv2 & 82.97 \\
VizWiz & 44.26 \\
rendered SST-2 & 100.00 \\
\bottomrule
\end{tabular}
    \caption{Evaluation results using IDEFICS 9B and demonstrations sampled with RICES using ground truth as similarity across twelve
vision-language datasets using 16 in-context demonstrations.}\label{tab:oracle}
\end{table}

\end{document}